\documentclass[runningheads]{llncs}

\usepackage{eccv}

\usepackage{eccvabbrv}

\usepackage{graphicx}
\usepackage{booktabs}

\usepackage[accsupp]{axessibility}  % Improves PDF readability for those with disabilities.

\usepackage{hyperref}

\usepackage{orcidlink}

\usepackage{amssymb}% http://ctan.org/pkg/amssymb
\usepackage{pifont}% http://ctan.org/pkg/pifont
\usepackage{amsmath,amssymb}
\usepackage{graphicx}
\usepackage{pifont}
\newcommand{\cxmark}{\ding{55}}
\usepackage{epsfig}
\usepackage{amsmath}
\usepackage{amssymb}
\usepackage{booktabs}
\usepackage{caption}
\usepackage{multirow}
\usepackage{algorithm2e}
\usepackage{xcolor}
\usepackage{graphicx}

\begin{document}

\title{HumanRefiner: Benchmarking Abnormal Human Generation and Refining with Coarse-to-fine Pose-Reversible Guidance} 

\titlerunning{HumanRefiner}

\author{Guian Fang \inst{1,}\thanks{Equal Contribution.} \and
Wenbiao Yan\inst{4,*} \and
Yuanfan Guo\inst{3,*} \and
Jianhua Han\inst{3} \and
Zutao Jiang\inst{2} \and
Hang Xu\inst{3} \and
Shengcai Liao\inst{5} \and
Xiaodan Liang\inst{1,2,}\thanks{Corresponding author}
}
\authorrunning{Fang and Yan et al.}

\institute{Sun Yat-sen University \and
Mohamed bin Zayed University of Artificial Intelligence
\\
\and
Huawei Noah's Ark Lab
\and
Xi'an Jiaotong University
\and
United Arab Emirates University
}

\maketitle

\vspace{-2mm}
\begin{abstract}
Text-to-image diffusion models have significantly advanced in conditional image generation. However, these models usually struggle with accurately rendering images featuring humans, resulting in distorted limbs and other anomalies. This issue primarily stems from the insufficient recognition and evaluation of limb qualities in diffusion models. To address this issue, we introduce \textbf{AbHuman}, the first large-scale synthesized human benchmark focusing on anatomical anomalies. This benchmark consists of 56K synthesized human images, each annotated with detailed, bounding-box level labels identifying 147K human anomalies in 18 different categories. Based on this, the recognition of human anomalies can be established, which in turn enhances image generation through traditional techniques such as negative prompting and guidance. To further boost the improvement, we propose \textbf{HumanRefiner}, a novel plug-and-play approach for the coarse-to-fine refinement of human anomalies in text-to-image generation. Specifically, HumanRefiner utilizes a self-diagnostic procedure to detect and correct issues related to both coarse-grained abnormal human poses and fine-grained anomaly levels, facilitating pose-reversible diffusion generation. Experimental results on the AbHuman benchmark demonstrate that HumanRefiner significantly reduces generative discrepancies, achieving a 2.9x improvement in limb quality compared to the state-of-the-art open-source generator SDXL and a 1.4x improvement over DALL-E 3 in human evaluations. Our data and code are available at \url{https://github.com/Enderfga/HumanRefiner}. 
\end{abstract}

\section{Introduction}
\label{sec:intro}
\begin{figure}[t]
    \vspace{-2mm}
    \centering
    \includegraphics[width=1\columnwidth]{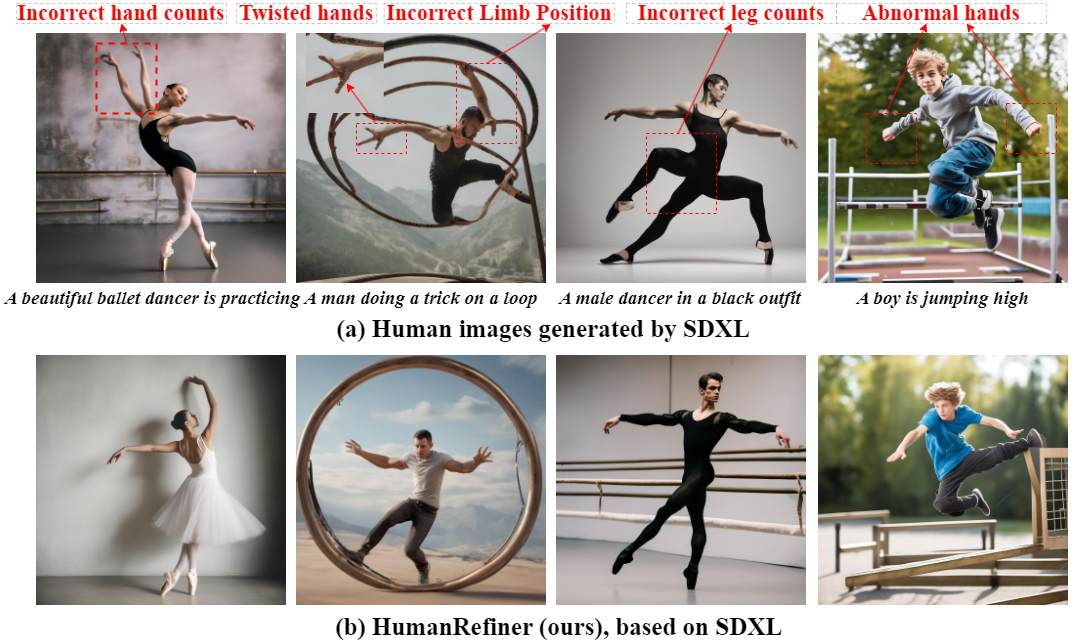}
    \caption{The synthesis of human images by SOTA text-to-image diffusion model SDXL~\cite{SDXL} still presents incorrect limb counts, twisted hands, and incorrect limb position problems. We propose \textit{HumanRefiner}, a coarse-to-fine self-diagnosis pose/anomaly-reversible generation pipeline built on the \textit{AbHuman} benchmark to eliminate the anomalies.}
    \label{fig:intro}
    \vspace{-5mm}
\end{figure}

Recent developments in text-to-image diffusion models, as exemplified by SDXL~\cite{SDXL}, DALL-E 2~\cite{DALLE_2}, and DeepFloyd \cite{SDXL, DALLE_2,shonenkov2023deepfloyd}, have marked significant breakthroughs in the generation of high-quality images.These models, benefiting from large-scale pre-training, are capable of generating images for various objects.  
However, generating complex and flexible structures, such as the human body, poses significant challenges for current text-to-image models. As illustrated in Figure~\ref{fig:intro}, common issues with these models include the generation of multiple limbs or abnormal limb details (e.g., irregular fingers). We attribute these challenges to the fact that upstream training typically focuses on aligning overall characteristics, like the human having hands and legs, while less attention is paid to detailed characteristics such as the number of fingers or hands. Consequently, models have limited control over the fine-grained generation of detailed body parts and lack the knowledge to distinguish between normal and abnormal human body characteristics. Though such knowledge could be introduced by extra pose condition as indicated in HumanSD~\cite{ju2023humansd}, Pose-ControlNet~\cite{controlnet}, T2I-Adapter~\cite{t2i_adapter}, these models are hard to use because of the requirement of providing extra input. Therefore, these models are not able to generalize to a wide variety of text-to-image generation applications. 

Therefore, this paper focuses on enhancing the text-to-image diffusion model's understanding of human anomalies and leveraging this knowledge to drive models for generating accurate and realistic human bodies. To address the above challenges, we introduce the first large-scale synthesized human benchmark for anatomical anomalies, named \textbf{AbHuman}. This benchmark comprises 56K synthesized human images, each meticulously annotated with detailed, bounding-box level labels that mark 147K instances of human limbs.
Instead of a single "abnormal body" category, these anomalies are categorized into 18 distinct finer groups, such as "abnormal/normal head", "abnormal/normal hand", "abnormal/normal foot", etc. 

With the proposed benchmark, it could be directly employed to provide knowledge about human anomalies by abnormal captions derived from the annotation, which can then be integrated into the diffusion model through the conventional negative prompting technique. Another possible way is to develop an Abnormal Scorer, which provides feedback on generated images to distinguish normal/abnormal images. Such a scorer enables quantitative assessment of anomalies in generated images and could provide guidance on anomalies for the text-to-image diffusion process.

However, these conventional approaches from our AbHuman benchmark do not fully leverage the detailed information available from bounding box-level annotations of human limb anomalies. Therefore, we propose \textbf{HumanRefiner}, a novel and adaptable method for addressing human anomalies that can be integrated with any text-to-image generation model. HumanRefiner goes beyond the typical abnormal guidance and negative prompting by implementing a coarse-to-fine, pose-reversible diffusion generation process. This includes a pose-guided generation for global refinement and anomaly detector-guided inpainting for fine-grained refinement. The detailed anomalies are identified using an abnormal detector train on AbHuman. Experimental results on the benchmark illustrate that HumanRefiner markedly reduces generative discrepancies, achieving a 2.9$\times$ improvement in limb quality compared to leading open-source generator SDXL and a 1.4$\times$ improvement over DALL-E 3 in human evaluations. In summary, our contributions are threefold:
\begin{itemize}
    \item  Introduction of AbHuman, the first large-scale benchmark of synthesized humans with anatomical anomalies, which encompasses 56K images, each annotated with detailed, bounding-box level human anomaly labels.
    \item We propose HumanRefiner, a novel plug-and-play approach for coarse-to-fine refinement of human anomalies via pose-reversible guidance.
    \item Comprehensive experimental results show that HumanRefiner significantly reduces generative discrepancies compared to SOTA text-to-image models. Human evaluation results indicate a 2.9$\times$ preference for our HumanRefiner in generating high-quality human images compared to SDXL and a 1.4$\times$ preference over DALL-E 3.
\end{itemize}

\section{Related Work}
\subsection{Text-to-Image/Human Generation}
\textbf{Text-To-Image Generation.} The goal of text-to-image synthesis is to generate realistic images based on given descriptions. Fueled by large-scale image-text datasets \cite{kakaobrain2022coyo_700m, schuhmann2022laion}, various training and inference techniques \cite{NEURIPS2020_4c5bcfec, song2020denoising, ho2022cascaded}, and advancements in scalability \cite{DALLE_2}, text-to-image models have made significant progress. Existing text-to-image models can be broadly categorized into three main approaches: those utilizing diffusion models \cite{balaji2022ediffi, DALLE_2, GLIDE, saharia2022photorealistic, gafni2022make, stablediffusion, Imagen_saharia2022photorealistic}, autoregressive models \cite{ramesh2021zero, yu2022scaling, chang2023muse}, and generative adversarial models \cite{sauer2023stylegan, kang2023scaling, zhang2022exploringGAN}. The latest models, such as DALL-E 3\cite{openai2023dalle3}, and SDXL\cite{SDXL}, have shown substantial improvements in image generation quality. However, the challenge of generating human-like images from text remains difficult due to the diversity of human body poses and insufficient alignment between image and text. Our work focuses on improving the quality of generated human images.

\noindent\textbf{Text-To-Human Generation.} Numerous models have made considerable efforts to enhance the generation quality of realistic human images. They have discovered that injecting additional conditions into diffusion models provides better guidance for human image generation. Models such as \cite{controlnet, t2i_adapter, huang2023composer} have incorporated additional training modules, introducing human pose conditions to finely control the generation of human images. Additionally, \cite{li2023gligen} has the capability not only to introduce key points but also to incorporate bounding box detection data. Building upon the introduction of additional poses, \cite{ju2023humansd} has implemented a strategy involving heatmap alignment to achieve higher-quality human images.

However, existing works dedicated to human generation face two main challenges: (1) the introduction of poses significantly restricts the diversity of generated human poses, and (2) pose conditions still exhibit limitations in addressing fine-grained details of human body parts, such as hands, feet, and facial features.

Differing from this line of research, HumanRefiner is designed to address the issue of human generation without the requirement of human-provided pose conditions.

\subsection{Human Generation Datasets.}
Currently, human datasets primarily consist of two types. The first focuses on authentic depictions of individuals in natural settings, such as human photos and images from human video data \cite{MSCOCO_lin2014microsoft, schuhmann2022laion, Pose2Seg_zhang2019pose2seg, HandDetec_narasimhaswamy2022whose, CrowdPose_li2019crowdpose, Posetrack_andriluka2018posetrack, Yutubepose_charles2016personalizing, open_pose_cao2017realtime}. These widely available datasets offer abundant human-pose pairs, with annotations limited to key points or bounding boxes. Given that these data are derived from real photos, the humans depicted are inherently natural, without limb anomalies. The second category involves artificial scenes featuring human-like figures in settings such as paintings, cartoons, and similar environments. The Human-Art dataset \cite{ju2023humanART} provides 50,000 images centered around humans, encompassing five natural scenes and fifteen artificial scenes, accompanied by precise pose and text annotations. None of these datasets provide large-scale synthesized images or potential limb abnormal annotations.

To address the challenge of limb anomalies in the text-driven human image generation task, we have developed a novel AbHuman dataset as a component of this research effort. This dataset serves as a valuable resource for effectively handling limb abnormalities within the context of generating human images from textual descriptions.

\section{The AbHuman Dataset}
\begin{figure}[t!]
    \centering
    \includegraphics[width=1\columnwidth]{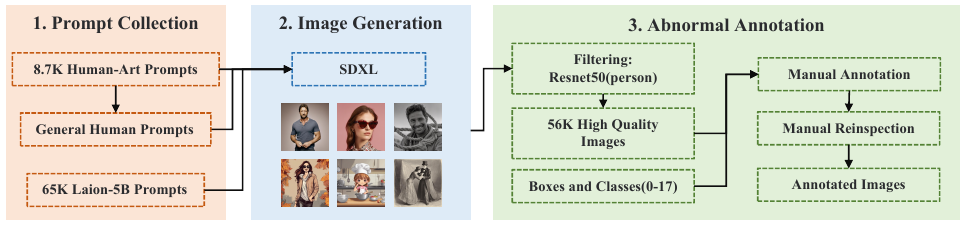}
    \caption{Data generation pipeline of our AbHuman dataset.}
    \vspace{-5mm}
    \label{fig_datasetpipeline}
\end{figure}
We provide a large-scale dataset of abnormal human images called AbHuman. Firstly, this is the first dataset to focus on limb abnormalities, as existing text-to-human algorithms are primarily based on the MS-COCO\cite{MSCOCO_lin2014microsoft} and LAION-5B\cite{schuhmann2022laion} datasets. Notably, models trained on real images often face challenges in accurately generating limb abnormalities in humans. To address this issue in general human image generation scenarios, we constructed a new benchmark. Secondly, this dataset collects a significant number of failure cases based on existing generative models. We observed that existing AI-generated human images often feature limb abnormalities, which we filtered, annotated, and analyzed. With the guidance of this dataset, the quality of generative models has been greatly improved.

\subsection{Data Collection}
We designed a standardized pipeline for abnormal human limb image collection, including prompt collection, image generation, and abnormal annotation, as shown in Fig. \ref{fig_datasetpipeline}.

\begin{itemize}
    \item \textbf{Prompt Collection: }Our prompts are derived from three sources: textual descriptions related to humans in the Laion-5B\cite{schuhmann2022laion}, textual descriptions from the Human-Art dataset\cite{ju2023humanART}, and prompts generated using ChatGPT 3.5\cite{GPT_brown2020language}. To extract text-generating human-related content from the Laion dataset, we employed keyword-based filtering to ensure the high quality of the textual prompts. This collection process resulted in 100,000 meticulously curated textual prompt entries. Subsequently, we utilized the collected textual content as the foundation of our dataset.
    \item \textbf{Image Generation: }With the rapid advancements in text-to-image models, the quality of human image generation has significantly improved. However, the occurrence of anomalies in generating human figures remains unresolved. Even state-of-the-art generation models like SDXL\cite{podell2023sdxl}, Midjourney, and DALL-E 3\cite{openai2023dalle3} encounter issues with limb abnormalities when generating human images. To enhance the quality of human generation using the best existing generation models as a baseline, we employed the SDXL model for generating our images.
    \item \textbf{Abnormal Annotation: }After collecting a large number of images, a comparative analysis revealed the presence of non-human images within the dataset. Initially, we applied a ResNet50-based classifier to filter all images labeled as "person". Following the removal of a subset of low-quality images, human annotators, who have been specifically trained, were assigned annotation tasks for the remaining 56,000 images. The annotated target detection dataset was subsequently subjected to review by two human evaluators, resulting in a curated AbHuman dataset comprising 56,000 images and 147,000 annotations for human limb detection. Details of the annotated categories are provided in the appendix.
    \item \textbf{Data Split: } The dataset is partitioned into training and testing splits in 4:1, with 52k and 13k image-text pairs, respectively. We further develop a testing split for human evaluation, where prompts are from HumanArt~\cite{ju2023humanART}, each describing humans performing difficult actions such as acrobatics, dancing, and drama.
\end{itemize}
    
As illustrated in Figure \ref{fig_detection_example}, the AbHuman dataset provides bounding boxes and classification information for abnormal humans. Unlike existing keypoint-based annotation methods, the bounding box annotation enables the identification of both the location and type of anomalies in limbs. By detecting and inpainting anomalies in human limbs, our dataset is more conducive to improving generation algorithms. These annotations were carried out by a professional team consisting of 10 data annotators and 2 data auditors, all extensively trained before annotation commencement to ensure high annotation quality and timely feedback.

\begin{figure}[tb]
  \centering
  \begin{minipage}{0.5\columnwidth}
    \centering
    \includegraphics[width=\linewidth]{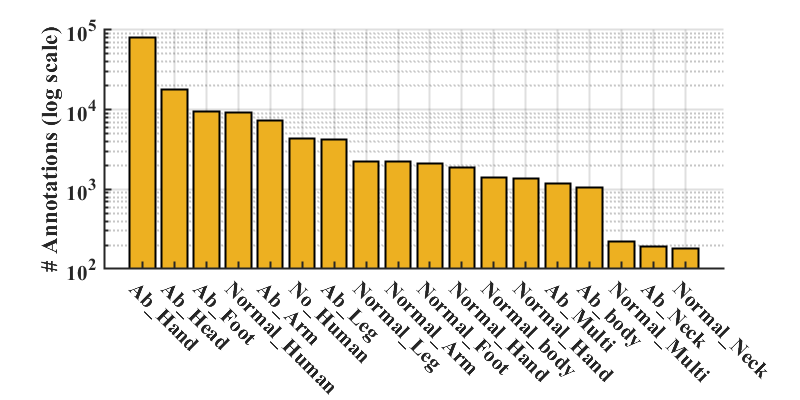}
    \caption{Ab-human annotations statistics.}
    \label{fig_bar_dataset}
  \end{minipage}%
  \begin{minipage}{0.5\columnwidth}
    \centering
    \includegraphics[width=\linewidth]{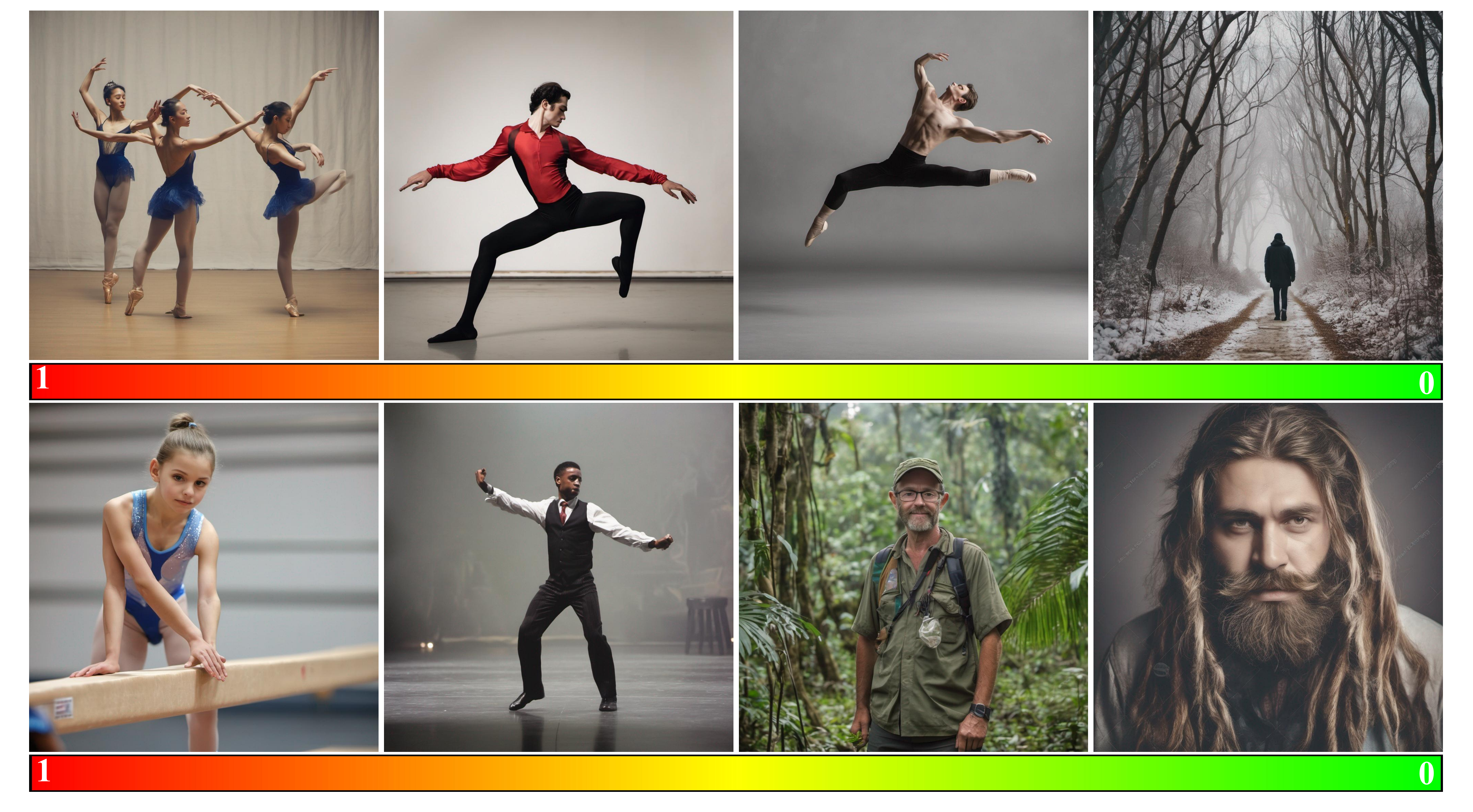}
    \caption{Visualization of abnormal scores. \textcolor{red}{red} indicates images with high abnormal scores while \textcolor{green}{green} indicates images with low abnormal scores.}
    \label{fig:abnormal-score-visualize}
  \end{minipage}
  % \caption{Two figures side by side}
  \label{fig:both}
  \vspace{-5mm}
\end{figure}

\subsection{Dataset Statistics and Analysis} \label{sec:3.2}
After reviewing a large number of generated samples, we conducted an analysis of human abnormalities and identified 18 distinct detection classes (\textit{e.g., Normal human, Abnormal Head, Abnormal Neck, Abnormal Body, Abnormal Arm, Abnormal Hand...}). To illustrate the class distribution of anomalous human samples generated by the SDXL model, we performed a statistical analysis of both anomalous humans and their corresponding annotations. As depicted in Figure \ref{fig_bar_dataset}, the model predominantly generates images with anomalous hands, aligning with the highest count of annotations for hand anomalies. Thus, it is evident that hand anomalies remain a challenge for text-to-image models. Additionally, the number of samples with anomalies in the head and foot regions is relatively higher compared to other parts of the human body.

Table~\ref{tab:addlabel} illustrates a comparison of datasets concerning human images. Compared to other human datasets\cite{HandDetec_narasimhaswamy2022whose, CrowdPose_li2019crowdpose, Multi_Person_PoseTrack_iqbal2017posetrack, Posetrack_andriluka2018posetrack, madhu2022enhancing, ju2023humanART} that solely annotate images of authentic and normal individuals, existing datasets have limited annotations for genuinely abnormal human limbs. On the contrary, the AbHuman dataset introduces anomalies in human-generated images and provides ample annotations for anomalous body postures. In the context of human-focused tasks such as generation and detection, AbHuman proves to be more suitable. Additionally, compared to datasets with relatively monotonous scenes, AbHuman encompasses a more diverse range of scenarios, enhancing its applicability in various contexts.

\begin{table}[htb]
\vspace{-2mm}
  \centering
  \scriptsize
  \caption{Comparison of human-centric datasets, including human generation and detection tasks. \textbf{AbHuman provides more granular annotation boxes and information to address the abnormality issues in the human generation.}}
    \begin{tabular}{lcccccc}
      \toprule
      Dataset & Images & Bbox  & Normal & Abnormal-Limb & General \\
      \midrule
      BodyHands\cite{HandDetec_narasimhaswamy2022whose} & 20K   & \checkmark     & \checkmark     & \cxmark     & \cxmark \\
      CrowdPose\cite{CrowdPose_li2019crowdpose} & 20K   & \checkmark     & \checkmark     & \cxmark     & \cxmark \\
      Multi-Person PoseTrack\cite{Multi_Person_PoseTrack_iqbal2017posetrack} & 23K   & \checkmark     & \checkmark     & \cxmark     & \cxmark \\
      PoseTrack\cite{Posetrack_andriluka2018posetrack} & 23K   & \checkmark     & \checkmark     & \cxmark     & \cxmark \\
      ClassArch\cite{madhu2022enhancing} & 1.5K  & \checkmark     & \checkmark     & \cxmark     & \cxmark \\
      Human-Art\cite{ju2023humanART} & 50K   & \checkmark     & \checkmark     & \cxmark     & \checkmark \\
      \midrule
      AbHuman (Ours) & 56K & \checkmark & \checkmark & \checkmark & \checkmark \\
      \bottomrule
    \end{tabular}
  \vspace{-5mm}
  \label{tab:addlabel}
\end{table}

\subsection{Abnormal Scorer} \label{sec:abnormal-scorer}

\begin{figure*}[t!]
    \centering
    \vspace{-2mm}
    \includegraphics[width=1\columnwidth]{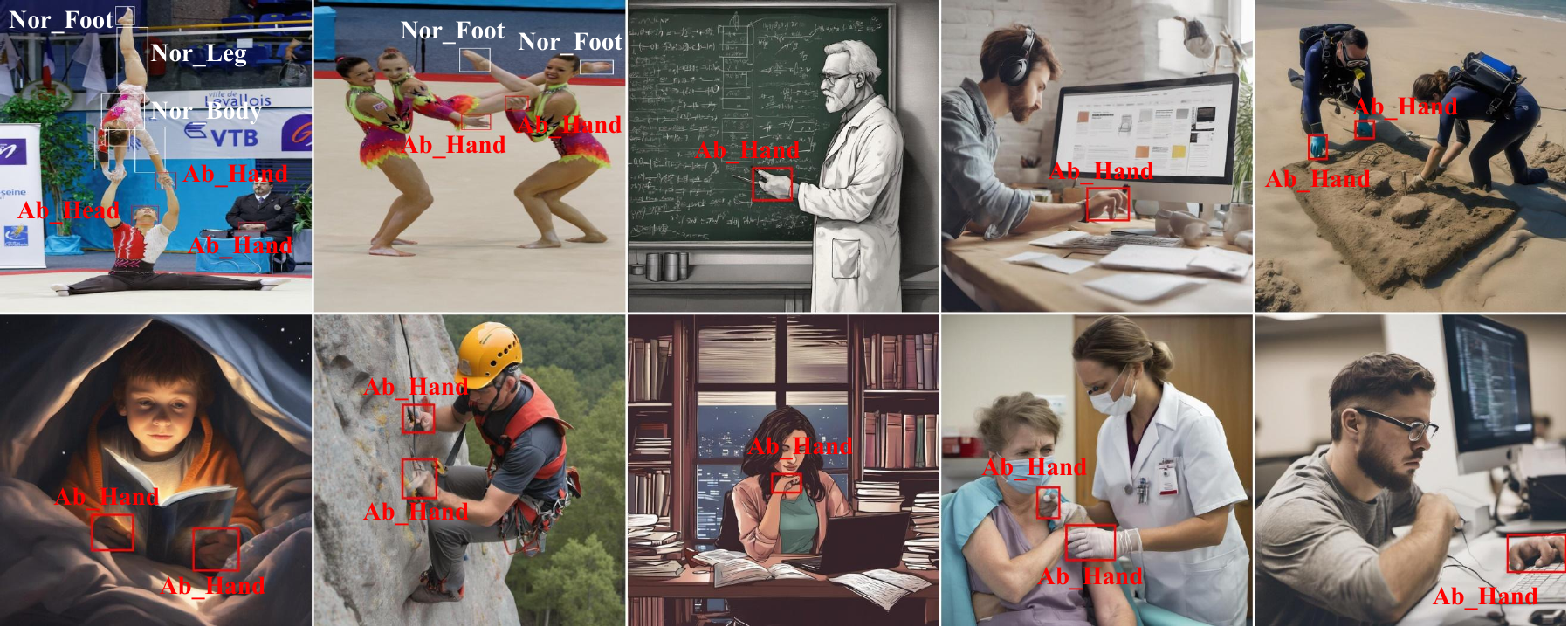}
    \caption{The visualization of limb detection in the test set using the fine-tuned AbHuman detection model. Red boxes and labels are used to annotate the abnormal limbs detected, and white boxes and labels are used to annotate the normal limbs detected.}
    % \xd{explain figures}}
    \label{fig_detection_example}
    \vspace{-5mm}
\end{figure*}

As indicated, the lack of a metric measuring the generated limb quality makes it difficult to (1) quantitatively compare different models; and (2) develop generators for the improvement of limb quality. Based on the proposed AbHuman Dataset, we present a quantitative metric of limb anomalies in a generated or real human image. To this end, inspired by the aesthetic scorer based on the Laion dataset~\cite{schuhmann2023laionaesthetics}, we train an Abnormal Scorer $S_\theta(\cdot)$.
For the model structure of Abnormal Scorer $S_\theta(\cdot)$, we adhere to the architecture utilized by Aesthetic Scorer~\cite{schuhmann2023laionaesthetics}. Building upon a pretrained CLIP ViT-Large with patch size of 14~\cite{CLIP} with a hidden dimension of 768, a  Multi-Layer Perceptron (MLP) is constructed, as depicted in Table~\ref{tab:model-architecture-abnormal-scorer}.

\begin{table}[htb]
    \centering
    \scriptsize
    \caption{Model architecture of MLP part of our abnormal scorer. Linear(A, B) means a linear layer with an input feature dimension of A and an output feature dimension of B. Dropout ($\alpha$) means a dropout layer with the probability of $\alpha$.}
    \begin{tabular}{c|cccc}
        \toprule
        Layer index & 0 & 1 &  2 &  3 \\
        \hline
        Layer & Linear (768, 1024) & Dropout (0.2) & Linear (1024, 128) & Dropout (0.2) \\
        \hline
        Layer index &  4&  5 & 6 & 7 \\
        \hline
        Layer & Linear (128, 64) & Dropout (0.1) & Linear (64, 16) & Linear (16, 1) \\
        \bottomrule
    \end{tabular}
    \label{tab:model-architecture-abnormal-scorer}
\end{table}

Given an image $x \in \mathcal{R}^{3 \times H \times W}$, the Abnormal Scorer assigns a score $S_\theta(x) \in [0,1]$ reflecting the severity of limb aberrations or distortions. Specifically, this model is architecturally based on a Multi-Layer Perceptron (MLP) and is built upon the foundation of the pre-trained CLIP Vision Transformer Large (ViT-L) model~\cite{CLIP}. This design choice allows for robust and nuanced assessments of abnormalities in image data. 

As illustrated in Figure~\ref{fig:abnormal-score-visualize}, high-scoring images are characterized by pronounced limb distortions or the absence of limbs, whereas images with lower scores correspond to either normal human depictions or non-human subjects. This scoring mechanism not only provides an empirical measure but also offers the potential for enhancing generative models. We provide a base solution in our HumanRefiner by offering Abnormal Guidance.

\vspace{-3mm}
\subsection{Abnormal Detector} \label{sec:detector}
\vspace{-2mm}
%% todo: add motivation?
To locate and detect abnormal limbs in human images for subsequent fine-grained refinement, we developed an abnormal limb detector. Specifically, we fine-tuned YOLOv8 \cite{yolo_redmon2016you} and RT-DETR \cite{lv2023detrs} on the training set of AbHuman for 100 epochs. Subsequently, we conduct corresponding detections on the test set, obtaining consistently stable results in the detection of abnormal limbs. Visualization results are provided in Figure~\ref{fig_detection_example}, showing the capability of identifying abnormal/normal regions and corresponding categories.

\begin{figure*}[t!]
    \centering
    \includegraphics[width=1.0\linewidth]{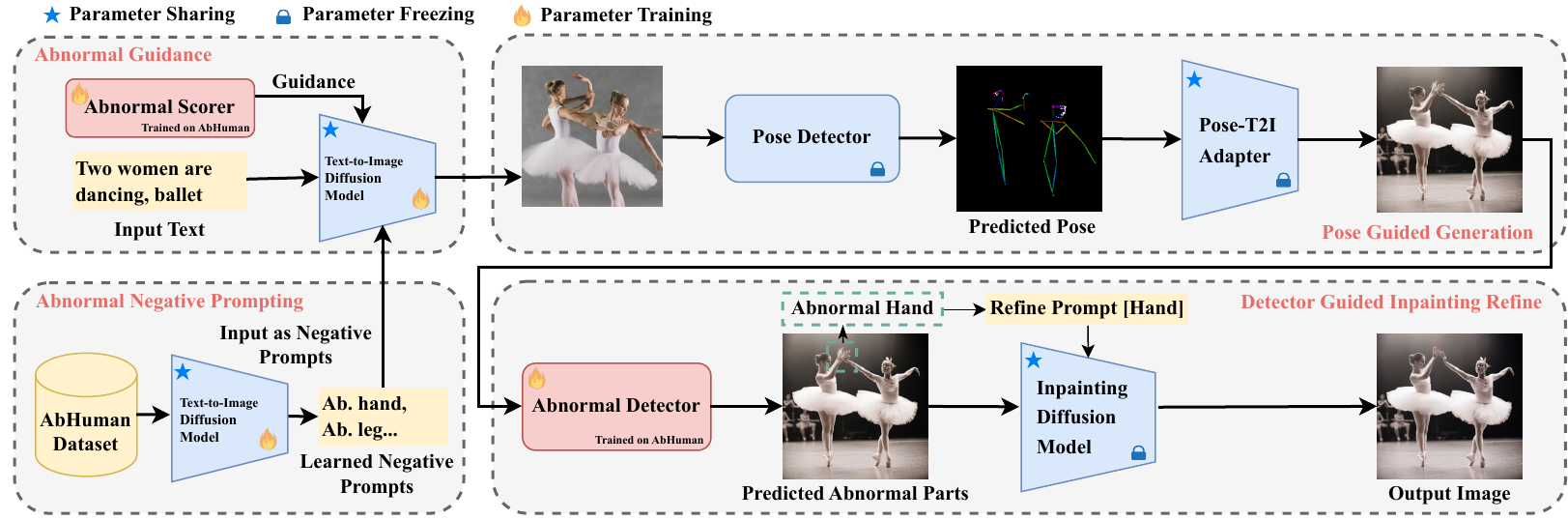}\vspace{-3mm}
    \caption{Overview of our HumanRefiner pipeline empowered by the proposed AbHuman benchmark. It consists of a Text-to-Image Diffusion Model trained with negative prompts on the AbHuman benchmark, which generates an initial human image with guidance from Abnormal Scorer, a \textbf{Coarse-to-Fine Refinement Process} where the generated image is globally refined by pose-guided generation and locally refined by detector-guided inpainting. For both the Pose-T2I Adapter~\cite{t2i_adapter} and the Inpainting Diffusion models, the Stable Diffusion parameters are adopted from the Text-to-Image model, which was refined through training in the Abnormal Guidance Stage.}
    \label{fig:humanrefiner-framework}
    \vspace{-5mm}
\end{figure*}

\section{HumanRefiner}

Based on the trained detector as illustrated in sec~\ref{sec:detector}, we can develop a flexible human image synthesis and refinement mechanism, \textbf{HumanRefiner}, which can be integrated into any pipeline for human image generation. As shown in Figure~\ref{fig:humanrefiner-framework}, in addition to the basic improvement on generation gained from negative prompting learned from Abnormal Guidance, HumanRefiner performs a coarse-to-fine pose-reversible refinement based on the initially generated images. An off-the-shelf pose detector~\cite{open_pose_cao2017realtime} is employed to inject pose guidance into the text-to-image diffusion model, which serves as conditions for coarse refinement of the initial generated image. Then, the Abnormal Detector identifies abnormals that are further rectified through an inpainting diffusion model, refining the remained abnormals at a fine-grained level.

The following sections demonstrate the basic control of human generation from negative prompting and abnormal guidance, followed by the coarse refinement module of pose-guided generation and the fine-grained refinement module of detector-guided inpainting.

\subsection{Negative Prompts }

The diffusion model is designed to generate data through a sequence of denoising autoencoders. The Latent Diffusion Model (LDM)~\cite{stablediffusion} conducts the diffusion process with a VAE encoder $\mathcal{E}$ and decoder $\mathcal{D}$. Given an image $x$, the forward diffusion process incrementally adds Gaussian noise $\epsilon$ to the latent $z_0=\mathcal{E}(x)$ following a T-step noise schedule $\{\alpha_t\}_{t=1}^T$ to produce a noisy latent $z_t$ where the noise level increases over timesteps $t$: 
\begin{equation}
    z_t = \sqrt{1 - \overline{\alpha}_t} z_0 + \sqrt{\overline{\alpha}_t} \epsilon,
\end{equation}
where $\epsilon \sim \mathcal{N}(0, I)$ and $\overline{\alpha}_t=\prod_{i=1}^t \alpha_i$. 
A text-to-image diffusion model learns a noise predictor $\epsilon_\theta(z_t, t, y)$ conditioned on text $y$ with the  objective:
\begin{equation}
    L = \mathbb{E}_{\mathcal{E}(x), y, \epsilon \sim \mathcal{N}(0,1),t}[ || \epsilon - \epsilon_\theta(z_t, t, y)||_2^2].
\end{equation}

The text condition $y$ ensures that the model's denoising is conditioned and controlled. Consequently, based on our AbHuman Benchmark, we can train the Latent Diffusion Model with the abnormal text conditions $y_a$(e.g., Ab. Hand, Ab. Feet...), making the model aware of the abnormal limb information. During the generation process, Classifier-Free Guidance~\cite{ho2022classifier} could be applied by using the abnormal text as a negative prompt to eliminate the generation of abnormal limbs:
\begin{equation}
\hat{\epsilon}_\theta(z_t, t, y) = \epsilon_\theta(z_t , t, y_a) + s \cdot (\epsilon_\theta(z_t, t, y) - \epsilon_\theta(z_t, t, y_a)),
\end{equation}
where $s$ is a scaling factor that modulates the influence of the condition.

\subsection{Abnormal Guidance}

In addition to adding information on human abnormals through negative prompting, we can introduce a general form of guidance to improve control over the denoising process. As shown in section~\ref{sec:abnormal-scorer}, the Abnormal Scorer $S_\phi(x)$ gives a score to an image that measures the severity of limb abnormals. This scorer can provide further guidance using a general Classifier Guidance~\cite{dhariwal2021diffusion}. Specifically, since $\epsilon_\theta(z_t, t, y)$ predicts the added noise, we can estimate the clean latent variable $z_0$ by: 
\begin{equation}
\hat{z_0} = \frac{z_t - \sqrt{\overline{\alpha}_t} \epsilon_\theta(z_t, t, y)}{\sqrt{1 - \overline{\alpha}_t}}.
\end{equation} 
After this, we can calculate an estimated abnormal score with 
$S_\phi(\mathcal{D}(\hat{z_0}))$. Consequently, the noise prediction guided by this method is:
\begin{equation}
\ddot{\epsilon}_\theta (z_t, t, y)  = \epsilon(z_t,t, y) + \mu \cdot \nabla S_\phi(\mathcal{D}(\hat{z_0})).
\end{equation}
We can combine classifier-free guidance through negative prompting with abnormal guidance easily by:
\begin{equation}
\begin{split}
    \hat{\epsilon}_t = & \epsilon_\theta(z_t, t, y_a) + s \cdot (\epsilon_\theta(z_t, t, y) - \epsilon_\theta(z_t , t, y_a)) \\
& + \mu \cdot \nabla S_\phi(\mathcal{D}(\hat{z_0})).  
\end{split}  
\end{equation}

%% formulate guidance

\subsection{Coarse-to-fine Pose-reversible Guidance}

The integration of negative prompting and abnormal guidance, empowered by our AbHuman Benchmark, offers robust steering for the human generation. In contrast to unguided approaches, our HumanRefiner now yields human images with more accurate poses initially. To further enhance the generation quality, we refine the generated image with pose-reversible guidance at both coarse and fine-grained levels.

\noindent\textbf{Pose-Guided Coarse Refine} The abnormal guided text-to-image diffusion model could be combined with off-the-shelf pose detector~\cite{open_pose_cao2017realtime} and T2I-Adapter~\cite{t2i_adapter} to further inject global human pose information for pose-guided refinement.

Specifically, for an initially generated image $\hat{x}$ from a guided text-to-image diffusion model, a pose keypoint map $M$ could be detected by a pose detector $M = P_\psi(\hat{x})$. The pose-guided generation could be expanded from the text-to-image diffusion model with T2I-Adapter~\cite{t2i_adapter} by injecting the pose condition information into intermediate encoder features:

\begin{equation}
    F_c = \mathcal{F}_{AD} (M),
\end{equation}
\vspace{-5.5mm}
\begin{equation}
    F_{enc} = F_{enc} + F_c,
\end{equation}

where $\mathcal{F}_{AD}$ is T2I-Adapter and $F_{enc}$ is the intermidate feature of the encoder part of $\epsilon_\theta(\cdot)$. Forming the pose conditioned denoising as $\epsilon_\theta(z_t, t, y, \mathcal{F}_{AD}(M))$, the advantage of abnormal guidance and negative prompts could also be integrated as:
\begin{equation}\label{eq9}
\begin{split}
    \hat{\epsilon}_t= & \epsilon_\theta(z_t, t, y_a) + s \cdot (\epsilon_\theta(z_t, t, y, \mathcal{F}_{AD}(M)) - \epsilon_\theta(z_t , t, y_a)) \\
& + \mu \cdot \nabla S_\phi(\mathcal{D}(\hat{z_0})).  
\end{split}  
\end{equation}

\begin{table*}[tb]
\centering
\caption{Abnormal score, CLIP Score and FID on AbHuman full test split (prompts from LAION) and hard test split (prompts from HumanArt).}
\vspace{-2mm}
\resizebox{1\columnwidth}{!}{
\begin{tabular}{c|ccc|ccc}
\hline
Dataset & \multicolumn{3}{c|}{LAION} & \multicolumn{3}{c}{HumanArt}\\
\hline
Model              & Abnormal Score $\downarrow$ & CLIP Score$\uparrow$& FID Score$\downarrow$ & Abnormal Score$\downarrow$ & CLIP Score$\uparrow$ &  FID Score$\downarrow$ \\ \hline
SSD-1B\cite{segmind_ssd1b}  & 0.612 & 34.02& 16.465 & 0.832 & {33.37} & 13.304 \\
PixArt-XL-2-1024-MS\cite{chen2023pixartalpha}  & 0.701 & 34.32& 19.604 & 0.807 & 32.68 & 15.452\\
DeepFloyd-IF\cite{shonenkov2023deepfloyd} & 0.595 & 32.72& 25.926 & 0.849 & 32.67 &21.512\\
LCM\cite{luo2023latent}  & 0.644 & 32.99& 25.175 & 0.831 & 32.40 & 21.179\\
SDXL\cite{podell2023sdxl} & 0.659 & 34.13& 30.024 & 0.857 & 32.40 & 20.796\\
Pose-ControlNet\cite{zhang2023adding} & 0.663 & 33.51& 15.368 & 0.838 & 31.88 & 13.047\\
Pose-T2I-Adapter\cite{mou2023t2i} & 0.624 &  34.79 & 15.658 & 0.807 & 32.72 &12.582\\
HumanSD\cite{ju2023humansd} & 0.661 & 33.78& 50.546 & 0.801 & 33.03 & 38.032\\
HumanRefiner (ours) & \textbf{0.590} & \textbf{34.85}& \textbf{13.634} & \textbf{0.778} & \textbf{33.90} &\textbf{9.145}\\
\bottomrule
\end{tabular}}
\label{tab:quantitative}
\vspace{-3mm}
\end{table*}

\noindent\textbf{Detector-Guided Inpainting Refine} 
The pose-guided generation refines the initial image in a global view, performing a coarse refinement. Besides, a fine-grained refinement could be established with our abnormal detector providing region-level abnormal information.

Specifically, for a refined image by applying diffusion denoise process in Eq.~(\ref{eq9}), $x_r = \mathcal{D}(\hat{z_0})$, abnormal detector $D_{Ab}(\cdot)$ outputs a set of $K$ bounding boxes and corresponding categories $\{B_i, C_i\}_{i=1}^K = D_{Ab}(x_r)$, where $B_i=\{x_i, y_i, w_i, h_i\}$ denotes the detected regions, $C_i \in \{0,..17\}$ denoting the detected categories defined in sec~\ref{sec:3.2}. For detected abnormal classes, the local region $B_i$ is inpainted with an input text conditioned on abnormal category $\hat{y}(C_i)$. The local region in the original image is then pasted by the inpainted output.

% \begin{table*}[htb]
% \centering
% \caption{Quantitative comparison on AbHuman, Laion and HumanArt dataset. For FID and Abnormal Score, lower is better. FID is calculated on Laion~\cite{schuhmann2022laion} and HumanArt~\cite{ju2023humanART}.}\vspace{-3mm}
% \begin{tabular}{c|cccc}

\begin{table*}[t!]
\centering
\scriptsize
\caption{Ablation Study on the proposed components, evaluated on the AbHuman test split (prompts from Laion) with Abnormal score.}
\vspace{-2mm}
\begin{tabular}{cccc|c}
\toprule
Negative Prompt & Abnormal Guidance & Detector Inpaint & Pose Guide &    Abnormal Score $\downarrow$ \\
\midrule
\cxmark & \cxmark & \cxmark & \cxmark & 0.659 \\
\checkmark & \cxmark & \cxmark & \cxmark & 0.614 \\
\checkmark & \checkmark & \cxmark & \cxmark & 0.605 \\
\checkmark & \checkmark & \checkmark & \cxmark & 0.601 \\
\checkmark & \checkmark & \checkmark & \checkmark & \textbf{0.590} \\
\bottomrule
\end{tabular}
\label{tab:ablation}
\vspace{-5mm}
\end{table*}

\section{Experiments}

\subsection{Implementation Details} \label{sec:implementation-details}

\noindent\textbf{Negative Prompt Fine-tuning}: we finetune SDXL~\cite{SDXL} released checkpoint at a resolution of 1024 x 1024 for 10 epochs with the learning rate of 1.0e-5 on the training split, with a batch size of 32 in total. The images are paired with the original text description followed by abnormal captions (e.g. Ab.hand, Ab.leg). More details can be found in the appendix.

\subsection{Quantitative Experiments} \label{sec:exps-quantitative}
\noindent\textbf{Evaluation Setup} We present quantitative comparison with baseline methods including SOTA text-to-image diffusion models SDXL~\cite{SDXL}, DeepFlyod-IF~\cite{shonenkov2023deepfloyd}, LCM~\cite{luo2023latent}. For methods require extra pose condition inputs such as HumanSD~\cite{ju2023humansd}, Pose-ControlNet~\cite{controlnet} and Pose-T2I-Adapter~\cite{t2i_adapter}, we first generate an initial image from SDXL~\cite{SDXL}, then generate pose condition from it with pose detector~\cite{openpose}, which operates similarly to our pose-guided generation. Test prompts are provided from the AbHuman full test split (prompts from LAION) and hard test split (prompts from HumanArt). We quantitatively evaluate the generated images with Abnormal Score, CLIP score and FID, respectively. Abnormal Score measures the severity of anomalies. CLIP score evaluates the image-text alignment by calculating the feature distance between input text and generated images. FID is calculated among the generated images and GT images from LAION or HumanArt.%, respectively. 

\noindent\textbf{Results} As indicated in Table~\ref{tab:quantitative}, HumanRefiner consistently achieves the best performance in terms of Abnormal score, FID, and CLIP score. Furthermore, when compared to Pose-ControlNet~\cite{controlnet} and Pose-T2I-Adapter~\cite{t2i_adapter} applied on SDXL, HumanRefiner demonstrates a significant improvement, indicating the effectiveness of HumanRefiner on integrating off-the-shelf methods such as pose detectors and T2I-adapter~\cite{t2i_adapter}. 

\label{sec:human-evaluation}
\begin{figure*}[htb]
    \vspace{-6mm}
    \centering
    \includegraphics[width=1\columnwidth]{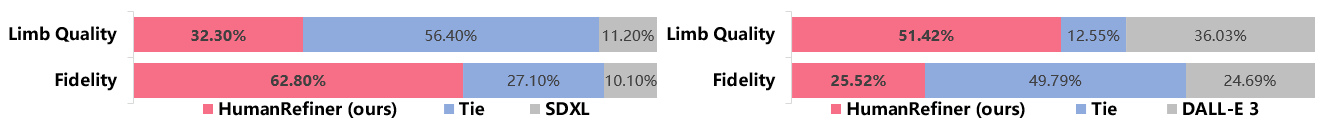}
    \caption{Human Evaluation on prompts from HumanArt. The percentage indicates the frequency of human preference. "Tie" means having a similar level of preference.}
    \label{fig:human evaluation}
    \vspace{-5mm}
\end{figure*}

\begin{figure*}[t!]
    \centering
\includegraphics[width=0.95\linewidth]{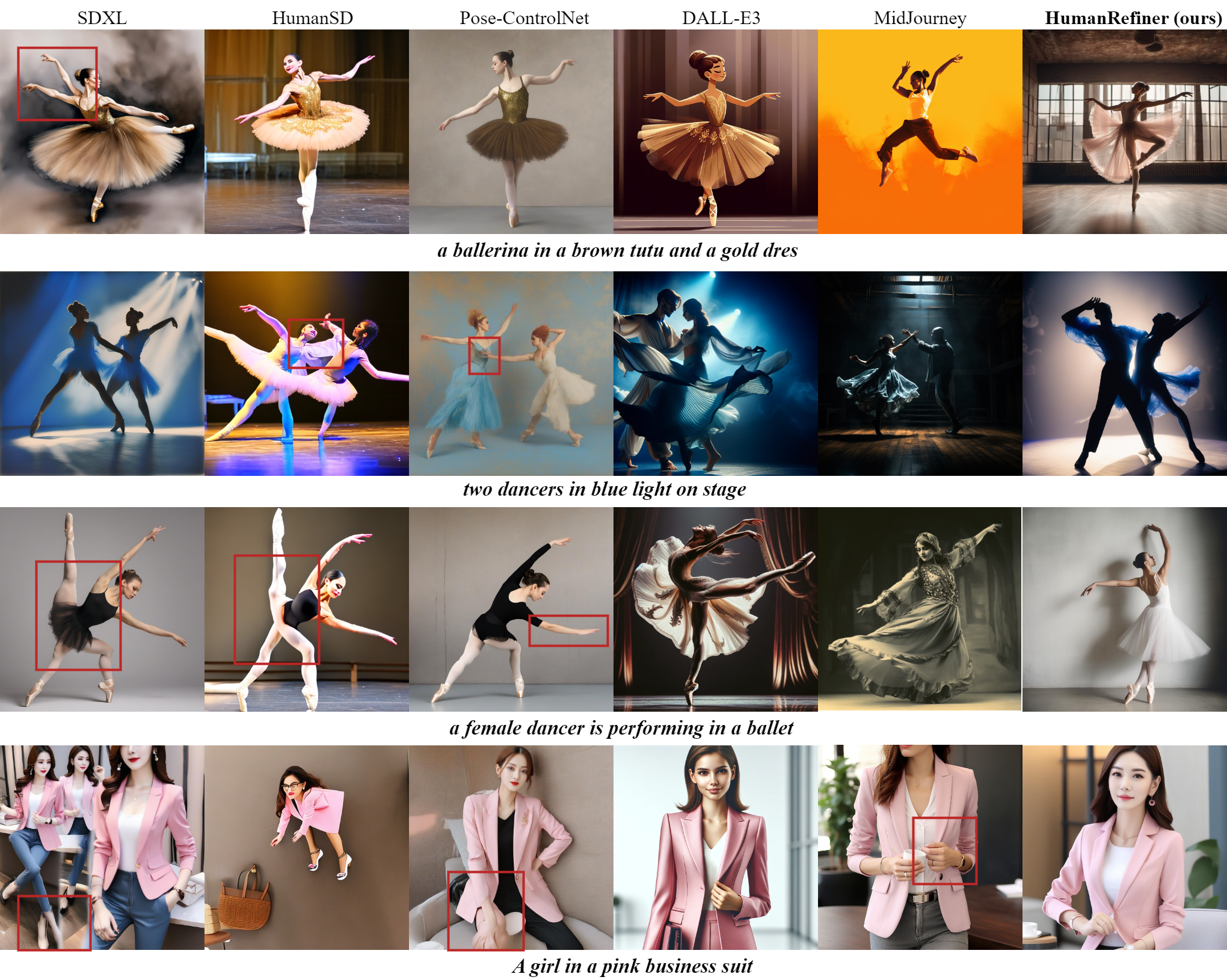}
    \vspace{-3mm}\caption{Qualitative comparison with baselines conducted on humans engaged in activity prompts.}
    \label{fig:visualize-compare}
    \vspace{-5mm}
\end{figure*}

\begin{figure}[tb]
    \centering
    \includegraphics[width=0.85\columnwidth]{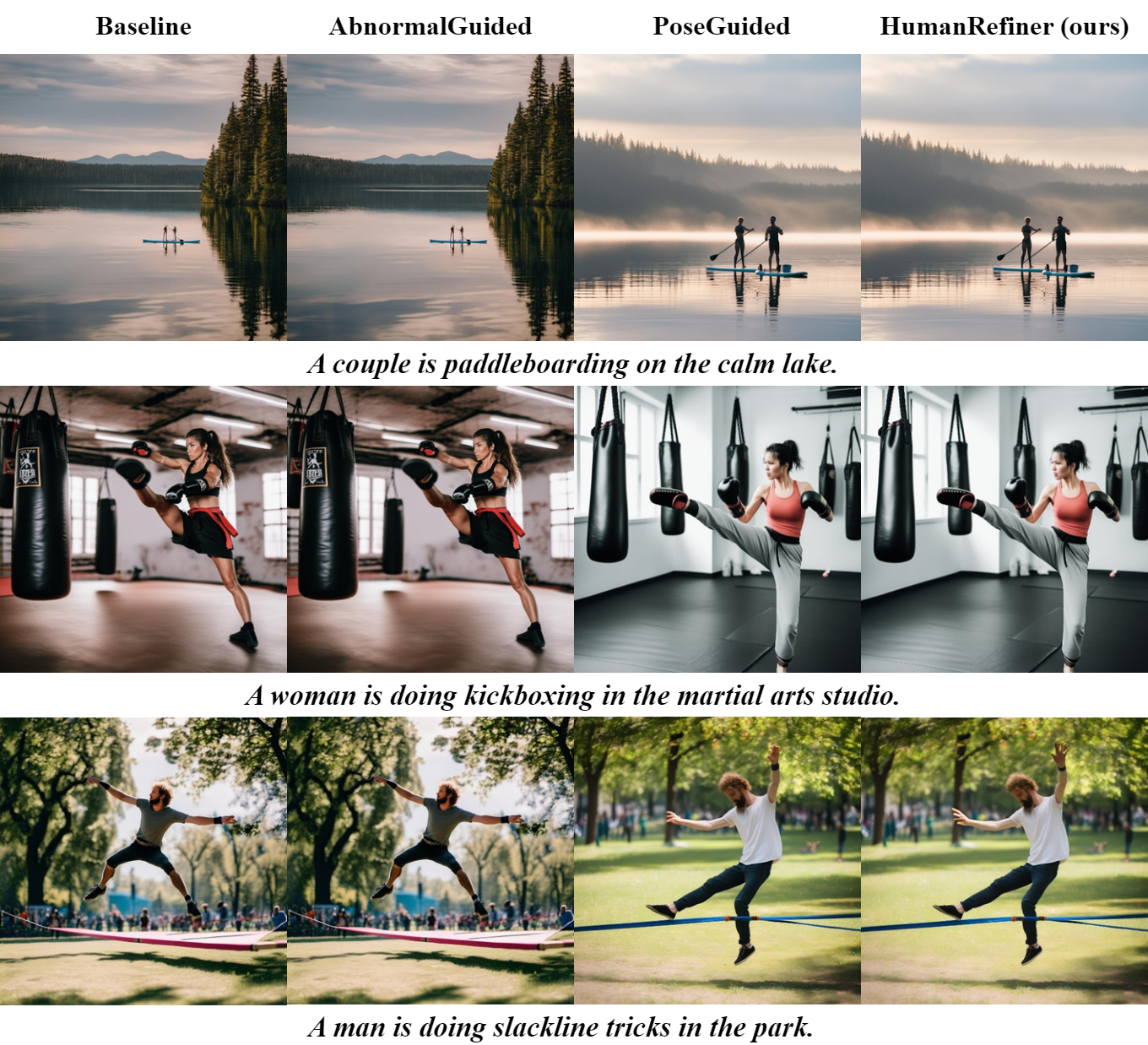}
    \caption{Visualization outcomes from ablation experiments.}
    \label{fig:ablation_examples}
    \vspace{-5mm}
\end{figure}

\noindent\textbf{Human Evaluation} We conduct a human preference study on generated images with prompts from HumanArt~\cite{ju2023humanART},  asking 38 users for a preference comparison between DALL-E 3~\cite{openai2023dalle3}, SDXL~\cite{SDXL}, and the proposed HumanRefiner. Each user is asked to compare the images from two aspects respectively, including fidelity and limb quality. Users could also choose "tie", meaning the two images are on the same level. Throughout the evaluation, users are unaware of which model the image is generated from. Figure~\ref{fig:human evaluation} shows that HumanRefiner surpasses DALL-E 3 and SDXL in both fidelity and limb quality for the human generation. For limb quality, HumanRefiner gets 2.9$\times$ preference over SDXL (32.3\% vs. 11.20\%), and 1.4$\times$ over DALL-E 3 %(25.52\% vs. 24.69\%)
(51.42\% vs. 36.03\%), indicating that HumanRefiner can generate high-quality human pose.

\noindent\textbf{Ablation Study}
We perform a component analysis to assess the contribution of each element utilizing the Abnormal Score as a metric for generation quality. As shown in Table~\ref{tab:ablation}, each component distinctly contributes to the Abnormal Score. For instance, the incorporation of the negative prompt learned from the AbHuman dataset significantly reduces the Abnormal Score from 0.659 to 0.614.

To offer a clearer presentation of HumanRefiner's corrective outcomes, we present several ablation experiment findings in Figure~\ref{fig:ablation_examples}. The \textbf{Baseline} represents the unaltered output of the SDXL model, while \textbf{AbnormalGuided} and \textbf{PoseGuided} illustrate results achieved through the application of distinct guiding modules. As depicted in the first row of figures, discernible enhancements are observed in the characters' outline. Moreover, the second and fourth rows showcase varying degrees of improvement in hand details. Notably, our AbnormalGuided module demonstrates corrective efficacy even in scenes where hand details typically appear blurry. Importantly, the ablation of our AbnormalGuided module mirrors an ablation of the dataset, underscoring the dataset's pivotal role in our approach.

\subsection{Qualitative Results}
We also present showcases in the challenging test split, featuring humans engaged in activities such as dancing or performing acrobatics. As illustrated in Figure~\ref{fig:visualize-compare}, HumanRefiner exhibits superior limb quality compared to the baseline methods. It accurately captures the human pose, represents natural limb attributes, and presents visually pleasing arrangements.

\section{Conclusion and Future work}

In this paper, we introduce AbHuman, a benchmark including fine-grained annotations of abnormal limbs. AbHuman comprises 56K generated human images along with 147K annotations for abnormal limbs at the bounding box level. Leveraging this benchmark, we construct HumanRefiner, a plug-and-play human refinement module that could be integrated into any text-to-image diffusion model. 
% \lsc{Looks like it is rather in the contrary: any existing T2I models can be integrated/applied in HumanRefiner.} 
Integraed with abnormal guidance, pose-reversible guidance, and detector-guided inpainting, HumanRefiner identifies abnormal limbs and removes them during the generation process, resulting in high-quality human image synthesis. Quantitative experimental results on the AbHuman test split, along with human evaluation results, demonstrate the superiority of HumanRefiner. We envision that the AbHuman Benchmark will facilitate further exploration of methods in the field of human image generation.

\noindent\textbf{Limitation and Future Work} While HumanRefiner is capable of generating high-quality human images, the process involves dual generation steps and inpainting, resulting in a limited inference speed. Furthermore, the AbHuman Dataset lacks broader scene annotations for normal human images. We would continue to improve the diversity of the AbHuman benchmark. We believe that richer annotations will further enhance the performance of HumanRefiner and foster the development of more effective models.

\section*{Acknowledgments}

This work was supported in part by National Science and Technology Major Project (2020AAA0109704), National Science and Technology Ministry Youth Talent Funding No. 2022WRQB002, Guangdong Outstanding Youth Fund (Grant No. 2021B1515020061), Mobility Grant Award under Grant No. M-0461, Shenzhen Science and Technology Program (Grant No. GJHZ20220913142600001), Nansha Key RD Program under Grant No.2022ZD014.

% \clearpage
\section*{Appendix}
\addcontentsline{toc}{section}{Appendix}
\appendix

\section{More Details on Dataset Collection}
In this section, we provide more details on dataset collection including the image generation process of our AbHuman dataset, the annotation process, and a detailed illustration of class definition with visualized examples.

\noindent\textbf{Image Generation}
Our dataset prompts are derived from three sources: Laion dataset, utilizing prompts extracted through keyword analysis (e.g., ``paper airplane in children's hands on a yellow background and blue sky on a cloudy day''); the second portion comprises textual descriptions from Human-Art (e.g., ``acrobatics, a woman performing aerial acrobatics on stage''); and the third part consists of language text generalized by GPT-3.5 based on scene descriptions from Human-Art (e.g., ``movie, a woman smoking a cigarette while sitting on a couch'').

After obtaining a substantial amount of text related to humans, we employed SDXL for image generation with a resolution of 1024*1024. Subsequently, we conducted a preliminary screening of the generated images. Using the Resnet50 classification model with the category 'person,' we identified images exhibiting a higher degree of relevance to humans.

\noindent\textbf{Class Definition} 
After observing a large number of generated images, we analyzed anomalies in existing human images. Eventually, we categorized the anomalies into 18 classes, annotated with indexes ranging from (0-17). Concerning the generation of images, we categorized them into ten classes: Normal People (0), Non-human (9), and Abnormal Annotations (1-8). To enhance the discriminative capabilities of the detector for both normal and abnormal instances, we introduced a set of authentic images. Notably, we annotated the normal parts as (10-17). The specific annotation results are illustrated in the accompanying Table~\ref{tab:your-label1} and \ref{tab:your-label2}.

\noindent\textbf{Annotation Process} 
An annotator employs labeling tools labelme~\cite{labelme} to annotate all objects within an image on a bounding box level, based on the category definitions and examples provided in Table~\ref{tab:your-label1} and \ref{tab:your-label2}. The categories labeled in the annotated image are then preserved as annotation text, as depicted in the Class \& explain column. This categorical text would be concatenated with the original descriptive text of the image, and such aggregated text is utilized for Negative prompt training.

\section{More Details on Abnormal Classifier and Abnormal Detector}

\subsection{Abnormal Detector}
We split the dataset into a training set and a test set with a ratio of 4:1. Subsequently, we fine-tuned YOLOv8~\cite{yolo_redmon2016you} and RT-DETR~\cite{lv2023detrs} to serve as our detectors. The model parameters used for fine-tuning YOLOv8 are detailed in the table~\ref{tab_detection_para}. The results obtained on the test set are summarized in the table~\ref{tab_detection_res}. Among them, YOLOv8n and YOLOv8x are two models in the YOLOv8 series, and YOLOv8x contains more layers and parameters. In our pursuit of enhanced detection performance, we incorporated YOLOv8x as the primary detector in our pipeline.

\begin{table}[h]
\scriptsize
    \centering
    \caption{The parameter Settings of the YOLOv8 model.}
    \begin{tabular}{c|ccccccccc|cccccccc}
    \hline
    \textbf{Parameter} & \textbf{nc} & \textbf{batch} & \textbf{patience} & \textbf{pre-trained} & \textbf{lr0} & \textbf{weight\_decay} & \textbf{NMS} & \textbf{iou} \\
     \textbf{Value} & 18 & 32 & 50 & true & 0.01 & 0.0005 & false &0.7  \\
     \hline
     \textbf{Parameter} &\textbf{epochs} & \textbf{images} & \textbf{workers} & \textbf{optimizer} & \textbf{momentum} & \textbf{augment} & \textbf{conf} & \textbf{max\_det} \\
      \textbf{Value} & 120 & 640 & 8 & auto & 0.937 & false & null & 300 \\
    \hline
    \end{tabular}
    \label{tab_detection_para}
\end{table}

\begin{table}[h]
\scriptsize
  \centering
  \caption{The detection results of abnormal limbs on the test set are presented for the AbHuman dataset.}
    \begin{tabular}{cccc}
    \toprule
    Model & Yolov8n & Yolov8x & RT-DETR \\
    \midrule
    mAP50 $\uparrow$ & 0.333 & \textbf{0.426} & 0.331 \\
    mAP50-95 $\uparrow$ & \textbf{0.282}  & 0.235  & 0.188 \\
    \bottomrule
    \end{tabular}%
  \label{tab_detection_res}%
\end{table}%

\subsection{Abnormal Scorer}

\noindent\textbf{Training Details} We finetune the Abnormal Scorer with a learning rate of 1.0e-4 using AdamW~\cite{loshchilov2017decoupled} optimizer with a batch size of 256 for 30epochs on the AbHuman training split. Images assigned with abnormal labels are positive samples while normal ones are negative samples. A sigmoid layer is followed by the last linear layer and binary cross-entropy is utilized as the objective function. 

\begin{figure*}[htb]
	\centering  %图片全局居中
	\subfloat[airtime, a young BMX rider in casual attire is catching air at a skatepark]{
		\includegraphics[width=\linewidth]{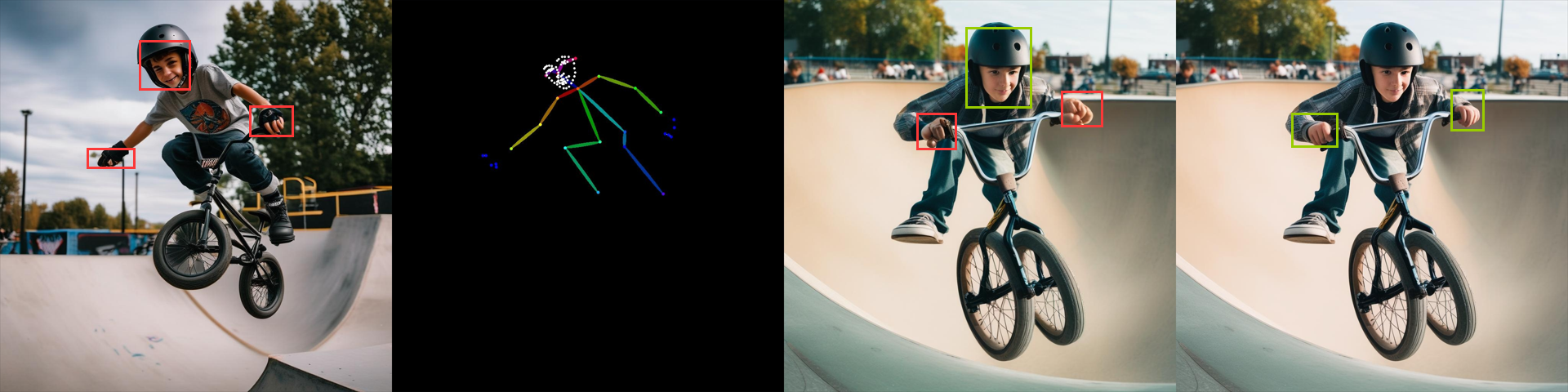}} \\
	\subfloat[combat, a player equipped for a paintball match is advancing through the forest]{
		\includegraphics[width=\linewidth]{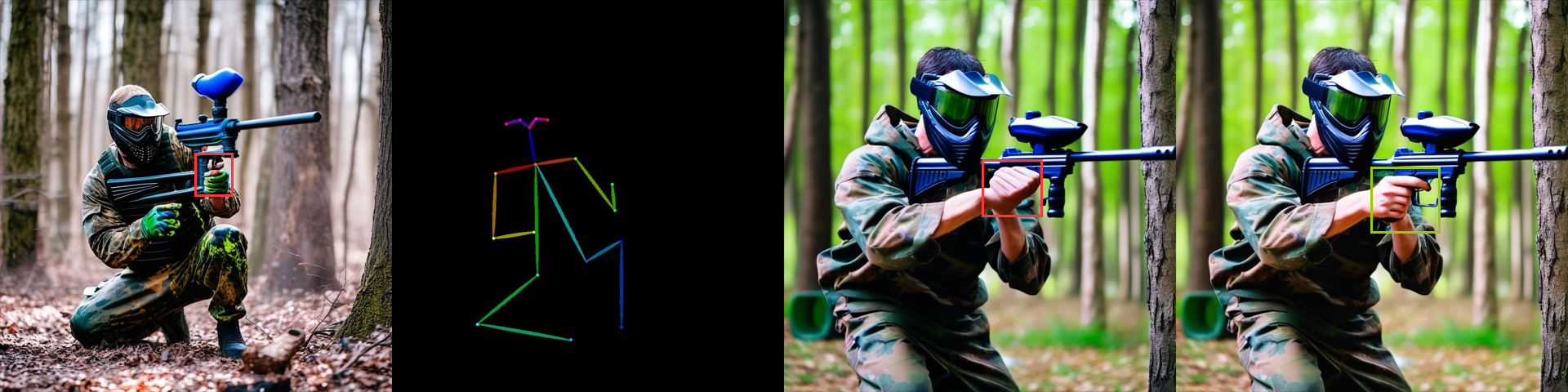}} \\
	\subfloat[asana, two individuals performing a yoga pose on a sandy beach at sunset]{
		\includegraphics[width=\linewidth]{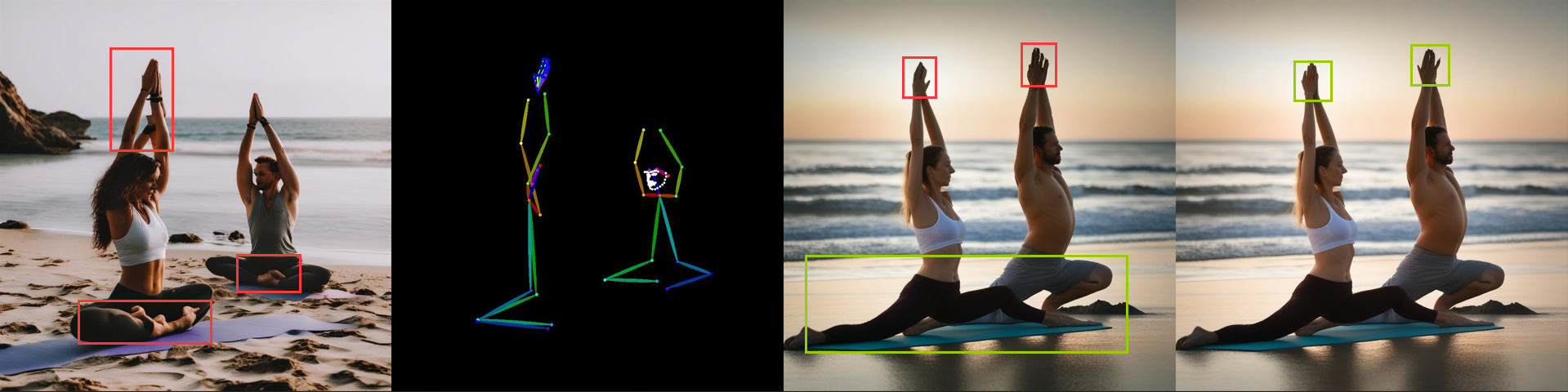}} \\
	\subfloat[dance, a ballet dancer in a white dress is standing in front of a wall]{
		\includegraphics[width=\linewidth]{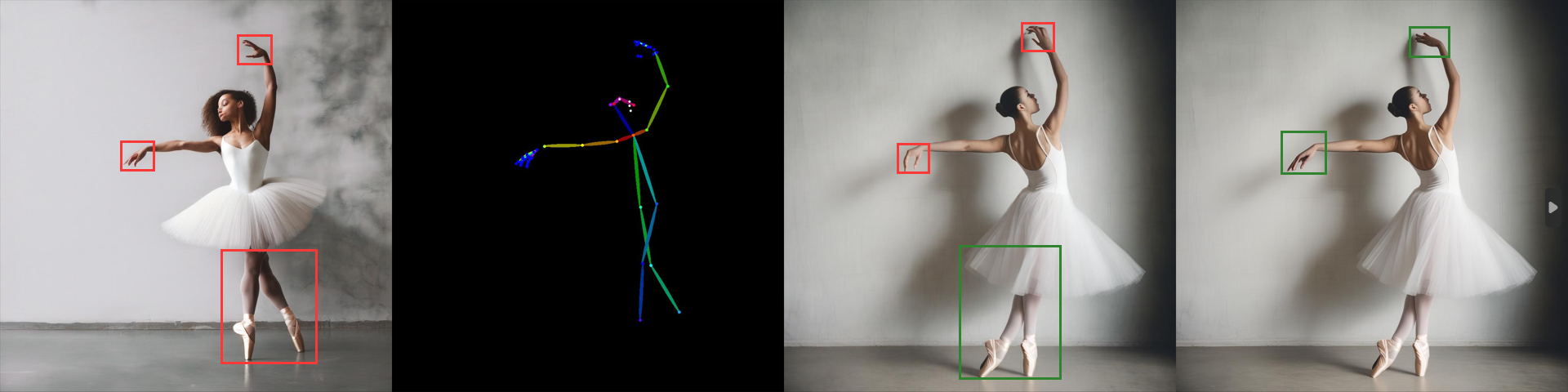}}
	\caption{Intermediate outputs of the HumanRefiner Pipeline from left to right: (1) Abnormal Guidance, (2) Pose Detection, (3) Pose-Guided Coarse Refinement, and (4) Inpainting Output. Highlight all the abnormal parts with a \textcolor{red}{red} box and mark the corresponding fixed normal parts with a \textcolor{green}{green} box. }
     \label{fig:visualize-examples-1}
\end{figure*}

\section{More Visualization of HumanRefiner }
We present the intermediate results of our HumanRefiner pipeline including the output of abnormal guidance, detected pose, pose-guided outputs, and the output of inpainting in Figure~\ref{fig:visualize-examples-1}.

\begin{table}[ht]
    \centering
        \caption{Annotations examples, class definition and corresponding textual explanation of AbHuman (Part 1).}
    \includegraphics[width=\textwidth]{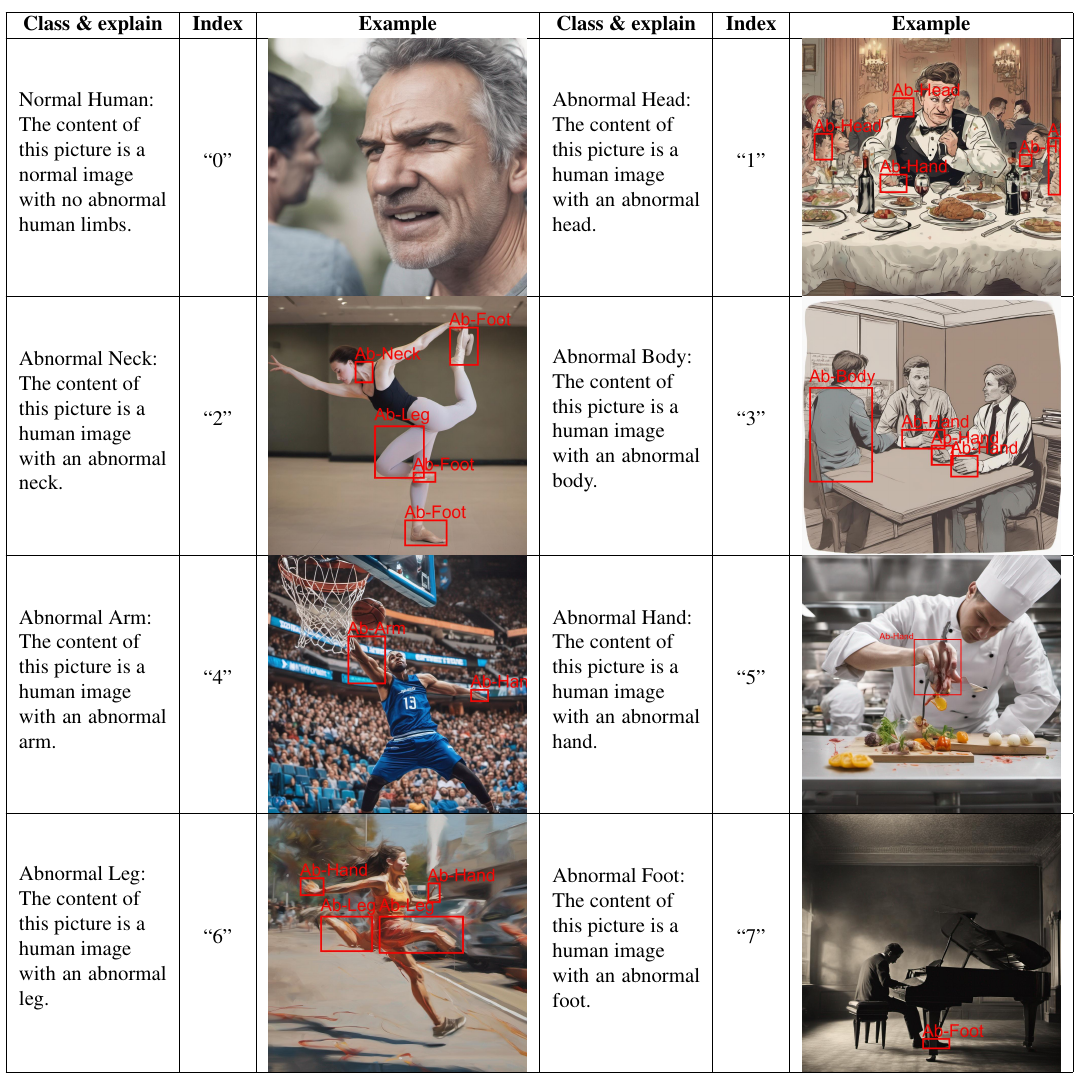}
    \label{tab:your-label1} 
\end{table}

\begin{table}[ht]
    \centering
     \caption{Continuation of annotations examples, class definition and corresponding textual explanation of AbHuman (Part 2).}
    \includegraphics[width=\textwidth]{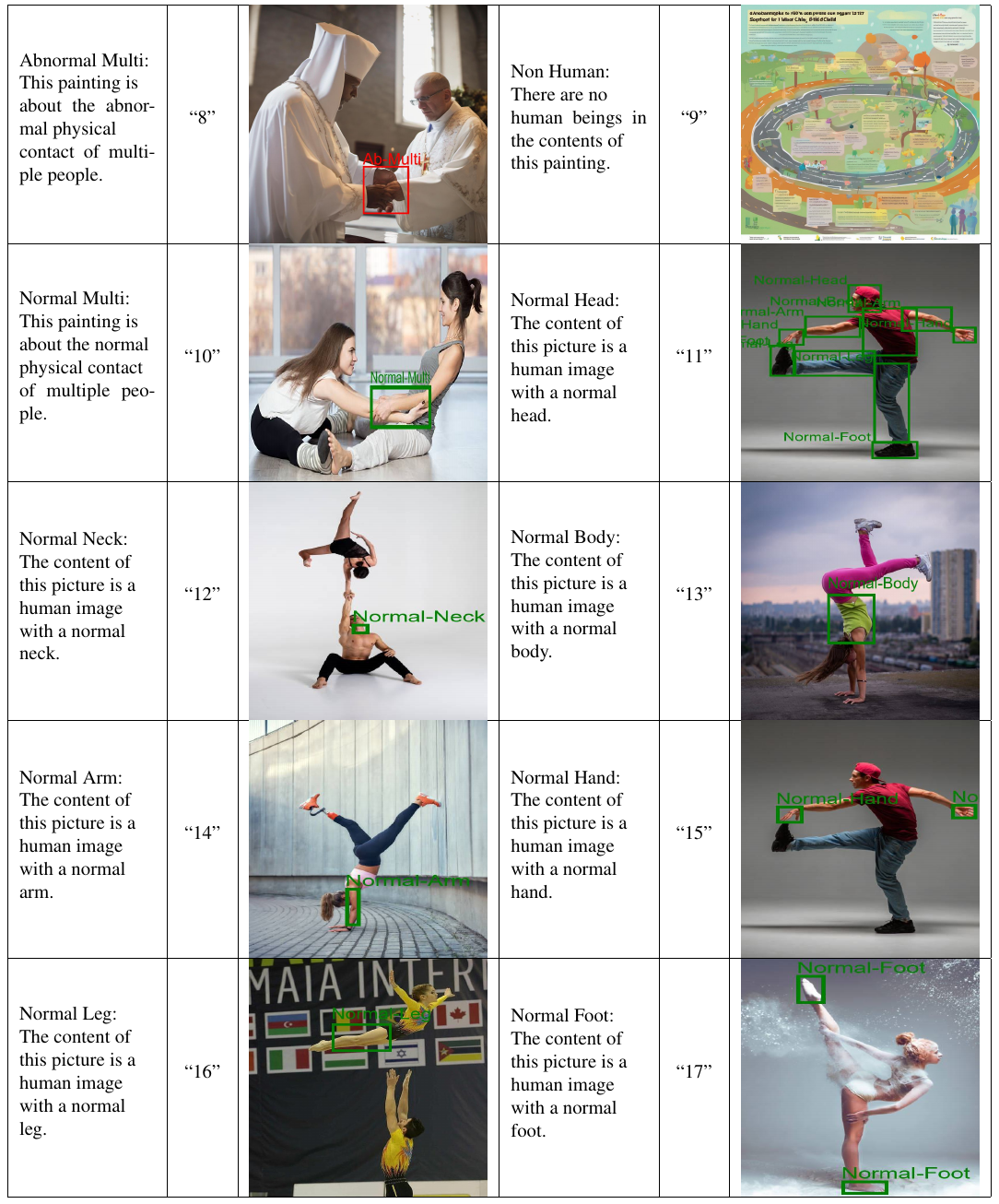}
    \label{tab:your-label2} 
\end{table}
%  % TODO REVIEW/FINAL: This \clearpage needs to be removed from both review and camera-ready versions.

\bibliographystyle{splncs04}
\bibliography{main}

\begin{thebibliography}{10}
\providecommand{\url}[1]{\texttt{#1}}
\providecommand{\urlprefix}{URL }
\providecommand{\doi}[1]{https://doi.org/#1}

\bibitem{Posetrack_andriluka2018posetrack}
Andriluka, M., Iqbal, U., Insafutdinov, E., Pishchulin, L., Milan, A., Gall, J., Schiele, B.: Posetrack: A benchmark for human pose estimation and tracking. In: Proceedings of the IEEE conference on computer vision and pattern recognition. pp. 5167--5176 (2018)

\bibitem{balaji2022ediffi}
Balaji, Y., Nah, S., Huang, X., Vahdat, A., Song, J., Kreis, K., Aittala, M., Aila, T., Laine, S., Catanzaro, B., et~al.: ediffi: Text-to-image diffusion models with an ensemble of expert denoisers. arXiv preprint arXiv:2211.01324  (2022)

\bibitem{GPT_brown2020language}
Brown, T., Mann, B., Ryder, N., Subbiah, M., Kaplan, J.D., Dhariwal, P., Neelakantan, A., Shyam, P., Sastry, G., Askell, A., et~al.: Language models are few-shot learners. Advances in neural information processing systems pp. 1877--1901 (2020)

\bibitem{kakaobrain2022coyo_700m}
Byeon, M., Park, B., Kim, H., Lee, S., Baek, W., Kim, S.: Coyo-700m: Image-text pair dataset. \url{https://github.com/kakaobrain/coyo-dataset} (2022)

\bibitem{openpose}
{Cao}, Z., {Hidalgo Martinez}, G., {Simon}, T., {Wei}, S., {Sheikh}, Y.A.: Openpose: Realtime multi-person 2d pose estimation using part affinity fields. IEEE Transactions on Pattern Analysis and Machine Intelligence  (2019)

\bibitem{open_pose_cao2017realtime}
Cao, Z., Simon, T., Wei, S.E., Sheikh, Y.: Realtime multi-person 2d pose estimation using part affinity fields. In: Proceedings of the IEEE conference on computer vision and pattern recognition. pp. 7291--7299 (2017)

\bibitem{chang2023muse}
Chang, H., Zhang, H., Barber, J., Maschinot, A., Lezama, J., Jiang, L., Yang, M.H., Murphy, K., Freeman, W.T., Rubinstein, M., et~al.: Muse: Text-to-image generation via masked generative transformers. arXiv preprint arXiv:2301.00704  (2023)

\bibitem{Yutubepose_charles2016personalizing}
Charles, J., Pfister, T., Magee, D., Hogg, D., Zisserman, A.: Personalizing human video pose estimation. In: Proceedings of the IEEE conference on computer vision and pattern recognition. pp. 3063--3072 (2016)

\bibitem{chen2023pixartalpha}
Chen, J., Yu, J., Ge, C., Yao, L., Xie, E., Wu, Y., Wang, Z., Kwok, J., Luo, P., Lu, H., Li, Z.: Pixart-$\alpha$: Fast training of diffusion transformer for photorealistic text-to-image synthesis (2023)

\bibitem{dhariwal2021diffusion}
Dhariwal, P., Nichol, A.: Diffusion models beat gans on image synthesis (2021)

\bibitem{gafni2022make}
Gafni, O., Polyak, A., Ashual, O., Sheynin, S., Parikh, D., Taigman, Y.: Make-a-scene: Scene-based text-to-image generation with human priors. In: European Conference on Computer Vision. pp. 89--106 (2022)

\bibitem{NEURIPS2020_4c5bcfec}
Ho, J., Jain, A., Abbeel, P.: Denoising diffusion probabilistic models. In: Advances in Neural Information Processing Systems. vol.~33, pp. 6840--6851 (2020)

\bibitem{ho2022cascaded}
Ho, J., Saharia, C., Chan, W., Fleet, D.J., Norouzi, M., Salimans, T.: Cascaded diffusion models for high fidelity image generation. The Journal of Machine Learning Research pp. 2249--2281 (2022)

\bibitem{ho2022classifier}
Ho, J., Salimans, T.: Classifier-free diffusion guidance. arXiv preprint arXiv:2207.12598  (2022)

\bibitem{huang2023composer}
Huang, L., Chen, D., Liu, Y., Shen, Y., Zhao, D., Zhou, J.: Composer: Creative and controllable image synthesis with composable conditions. arXiv preprint arXiv:2302.09778  (2023)

\bibitem{Multi_Person_PoseTrack_iqbal2017posetrack}
Iqbal, U., Milan, A., Gall, J.: Posetrack: Joint multi-person pose estimation and tracking. In: Proceedings of the IEEE Conference on Computer Vision and Pattern Recognition. pp. 2011--2020 (2017)

\bibitem{ju2023humanART}
Ju, X., Zeng, A., Wang, J., Xu, Q., Zhang, L.: Human-art: A versatile human-centric dataset bridging natural and artificial scenes. In: Proceedings of the IEEE/CVF Conference on Computer Vision and Pattern Recognition (2023)

\bibitem{ju2023humansd}
Ju, X., Zeng, A., Zhao, C., Wang, J., Zhang, L., Xu, Q.: Human{SD}: A native skeleton-guided diffusion model for human image generation  (2023)

\bibitem{kang2023scaling}
Kang, M., Zhu, J.Y., Zhang, R., Park, J., Shechtman, E., Paris, S., Park, T.: Scaling up gans for text-to-image synthesis. In: Proceedings of the IEEE/CVF Conference on Computer Vision and Pattern Recognition. pp. 10124--10134 (2023)

\bibitem{CrowdPose_li2019crowdpose}
Li, J., Wang, C., Zhu, H., Mao, Y., Fang, H.S., Lu, C.: Crowdpose: Efficient crowded scenes pose estimation and a new benchmark. In: Proceedings of the IEEE/CVF conference on computer vision and pattern recognition. pp. 10863--10872 (2019)

\bibitem{li2023gligen}
Li, Y., Liu, H., Wu, Q., Mu, F., Yang, J., Gao, J., Li, C., Lee, Y.J.: Gligen: Open-set grounded text-to-image generation  (2023)

\bibitem{MSCOCO_lin2014microsoft}
Lin, T.Y., Maire, M., Belongie, S., Hays, J., Perona, P., Ramanan, D., Doll{\'a}r, P., Zitnick, C.L.: Microsoft coco: Common objects in context. In: Computer Vision--ECCV 2014: 13th European Conference, Zurich, Switzerland, September 6-12, 2014, Proceedings, Part V 13. pp. 740--755. Springer (2014)

\bibitem{loshchilov2017decoupled}
Loshchilov, I., Hutter, F.: Decoupled weight decay regularization. arXiv preprint arXiv:1711.05101  (2017)

\bibitem{luo2023latent}
Luo, S., Tan, Y., Huang, L., Li, J., Zhao, H.: Latent consistency models: Synthesizing high-resolution images with few-step inference (2023)

\bibitem{lv2023detrs}
Lv, W., Xu, S., Zhao, Y., Wang, G., Wei, J., Cui, C., Du, Y., Dang, Q., Liu, Y.: Detrs beat yolos on real-time object detection (2023)

\bibitem{madhu2022enhancing}
Madhu, P., Villar-Corrales, A., Kosti, R., Bendschus, T., Reinhardt, C., Bell, P., Maier, A., Christlein, V.: Enhancing human pose estimation in ancient vase paintings via perceptually-grounded style transfer learning. ACM Journal on Computing and Cultural Heritage  \textbf{16}(1),  1--17 (2022)

\bibitem{mou2023t2i}
Mou, C., Wang, X., Xie, L., Wu, Y., Zhang, J., Qi, Z., Shan, Y., Qie, X.: T2i-adapter: Learning adapters to dig out more controllable ability for text-to-image diffusion models. arXiv preprint arXiv:2302.08453  (2023)

\bibitem{t2i_adapter}
Mou, C., Wang, X., Xie, L., Zhang, J., Qi, Z., Shan, Y., Qie, X.: T2i-adapter: Learning adapters to dig out more controllable ability for text-to-image diffusion models. arXiv preprint arXiv:2302.08453  (2023)

\bibitem{HandDetec_narasimhaswamy2022whose}
Narasimhaswamy, S., Nguyen, T., Huang, M., Hoai, M.: Whose hands are these? hand detection and hand-body association in the wild. In: Proceedings of the IEEE/CVF Conference on Computer Vision and Pattern Recognition. pp. 4889--4899 (2022)

\bibitem{GLIDE}
Nichol, A., Dhariwal, P., Ramesh, A., Shyam, P., Mishkin, P., McGrew, B., Sutskever, I., Chen, M.: Glide: Towards photorealistic image generation and editing with text-guided diffusion models. arXiv preprint arXiv:2112.10741  (2021)

\bibitem{openai2023dalle3}
OpenAI: Improving image generation with better captions (2023), \url{https://cdn.openai.com/papers/dall-e-3.pdf}, online; accessed 16-November-2023

\bibitem{SDXL}
Podell, D., English, Z., Lacey, K., Blattmann, A., Dockhorn, T., Müller, J., Penna, J., Rombach, R.: Sdxl: Improving latent diffusion models for high-resolution image synthesis (2023)

\bibitem{podell2023sdxl}
Podell, D., English, Z., Lacey, K., Blattmann, A., Dockhorn, T., Müller, J., Penna, J., Rombach, R.: Sdxl: Improving latent diffusion models for high-resolution image synthesis (2023)

\bibitem{CLIP}
Radford, A., Kim, J.W., Hallacy, C., Ramesh, A., Goh, G., Agarwal, S., Sastry, G., Askell, A., Mishkin, P., Clark, J., et~al.: Learning transferable visual models from natural language supervision. In: International conference on machine learning. pp. 8748--8763. PMLR (2021)

\bibitem{DALLE_2}
Ramesh, A., Dhariwal, P., Nichol, A., Chu, C., Chen, M.: Hierarchical text-conditional image generation with clip latents. arXiv preprint arXiv:2204.06125  \textbf{1}(2), ~3 (2022)

\bibitem{ramesh2021zero}
Ramesh, A., Pavlov, M., Goh, G., Gray, S., Voss, C., Radford, A., Chen, M., Sutskever, I.: Zero-shot text-to-image generation. In: International Conference on Machine Learning. pp. 8821--8831. PMLR (2021)

\bibitem{yolo_redmon2016you}
Redmon, J., Divvala, S., Girshick, R., Farhadi, A.: You only look once: Unified, real-time object detection. In: Proceedings of the IEEE conference on computer vision and pattern recognition. pp. 779--788 (2016)

\bibitem{stablediffusion}
Rombach, R., Blattmann, A., Lorenz, D., Esser, P., Ommer, B.: High-resolution image synthesis with latent diffusion models. In: Proceedings of the IEEE/CVF conference on computer vision and pattern recognition. pp. 10684--10695 (2022)

\bibitem{saharia2022photorealistic}
Saharia, C., Chan, W., Saxena, S., Li, L., Whang, J., Denton, E.L., Ghasemipour, K., Gontijo~Lopes, R., Karagol~Ayan, B., Salimans, T., et~al.: Photorealistic text-to-image diffusion models with deep language understanding. Advances in Neural Information Processing Systems  \textbf{35},  36479--36494 (2022)

\bibitem{Imagen_saharia2022photorealistic}
Saharia, C., Chan, W., Saxena, S., Li, L., Whang, J., Denton, E.L., Ghasemipour, K., Gontijo~Lopes, R., Karagol~Ayan, B., Salimans, T., et~al.: Photorealistic text-to-image diffusion models with deep language understanding. Advances in Neural Information Processing Systems  \textbf{35},  36479--36494 (2022)

\bibitem{sauer2023stylegan}
Sauer, A., Karras, T., Laine, S., Geiger, A., Aila, T.: Stylegan-t: Unlocking the power of gans for fast large-scale text-to-image synthesis. arXiv preprint arXiv:2301.09515  (2023)

\bibitem{schuhmann2023laionaesthetics}
Schuhmann, C.: Laion-aesthetics predictor v2 (2023), \url{https://laion.ai/blog/laion-aesthetics/}, online; accessed 16-November-2023

\bibitem{schuhmann2022laion}
Schuhmann, C., Beaumont, R., Gordon, C.W., Wightman, R., Coombes, T., Katta, A., Mullis, C., Schramowski, P., Kundurthy, S.R., Crowson, K., et~al.: Laion-5b: An open large-scale dataset for training next generation image-text models  (2022)

\bibitem{segmind_ssd1b}
Segmind: Announcing ssd-1b: A leap in efficient t2i generation (2023), \url{https://blog.segmind.com/introducing-segmind-ssd-1b/}, online; accessed 16-November-2023

\bibitem{shonenkov2023deepfloyd}
Shonenkov, A., Konstantinov, M., Bakshandaeva, D., Schuhmann, C., Ivanova, K., Klokova, N.: Deepfloyd if: A powerful text-to-image model that can smartly integrate text into images (2023), \url{https://www.deepfloyd.ai/deepfloyd-if}, online; accessed 16-November-2023

\bibitem{song2020denoising}
Song, J., Meng, C., Ermon, S.: Denoising diffusion implicit models. arXiv preprint arXiv:2010.02502  (2020)

\bibitem{labelme}
Wada, K.: labelme: Image polygonal annotation with python. \url{https://github.com/wkentaro/labelme} (2018)

\bibitem{yu2022scaling}
Yu, J., Xu, Y., Koh, J.Y., Luong, T., Baid, G., Wang, Z., Vasudevan, V., Ku, A., Yang, Y., Ayan, B.K., et~al.: Scaling autoregressive models for content-rich text-to-image generation. arXiv preprint arXiv:2206.10789  \textbf{2}(3), ~5 (2022)

\bibitem{controlnet}
Zhang, L., Rao, A., Agrawala, M.: Adding conditional control to text-to-image diffusion models. In: Proceedings of the IEEE/CVF International Conference on Computer Vision. pp. 3836--3847 (2023)

\bibitem{zhang2023adding}
Zhang, L., Rao, A., Agrawala, M.: Adding conditional control to text-to-image diffusion models (2023)

\bibitem{zhang2022exploringGAN}
Zhang, P., Yang, L., Lai, J.H., Xie, X.: Exploring dual-task correlation for pose guided person image generation. In: Proceedings of the IEEE/CVF Conference on Computer Vision and Pattern Recognition. pp. 7713--7722 (2022)

\bibitem{Pose2Seg_zhang2019pose2seg}
Zhang, S.H., Li, R., Dong, X., Rosin, P., Cai, Z., Han, X., Yang, D., Huang, H., Hu, S.M.: Pose2seg: Detection free human instance segmentation. In: Proceedings of the IEEE/CVF conference on computer vision and pattern recognition. pp. 889--898 (2019)

\end{thebibliography}
\end{document}